\documentclass[]{fairmeta_gr2}

\usepackage{graphicx}
\usepackage{booktabs}
\usepackage{tabularx}
\usepackage{multirow}
\usepackage{makecell}
\usepackage{array}
\usepackage{amsmath}
\usepackage{amssymb}
\usepackage{algorithm}
\usepackage{algpseudocode}
\usepackage{tikz}
\usetikzlibrary{arrows.meta}
\usepackage{placeins}
\graphicspath{{figures/}}
\newcommand{\ourmethod}{Self-Guided TTT}
\usepackage{xcolor}

\title{Self-Guided Test-Time Training for Long-Context LLMs}

\author[1,2]{Xinyu Zhu}
\author[1, \dagger]{Zhe Xu}
\author[1]{Xiaohan Wei}
\author[1]{Yunchen Pu}
\author[1]{Fei Tian}
\author[1]{Chonglin Sun}
\author[1]{Kaushik Rangadurai}
\author[1]{Hua Zhi}
\author[1]{Frank Shyu}
\author[1]{Sandeep Pandey}
\author[1]{Luke Simon}
\author[2,\ddagger]{Yu Meng}
\author[1,\ddagger]{Xi Liu}
\affiliation[1]{Meta AI}
\affiliation[2]{University of Virginia}
\metadata[Correspondence]{Yu Meng (\email{yumeng5@virginia.edu}) and Xi Liu (\email{xliu1@meta.com})}
\metadata[Date]{July 10, 2026}

\contribution[\dagger]{Execution Lead}
\contribution[\ddagger]{Joint corresponding author}

\abstract{Long-context processing has become increasingly important for large language models (LLMs), but simply extending the context window does not guarantee effective utilization of long inputs. As input length grows, accuracy often degrades, indicating that models still struggle to identify and use the evidence most relevant to a question. A promising way to improve long-context utilization is test-time training (TTT), which treats the test context as a training example for instance-specific parameter adaptation.
However, applying TTT to the entire long context is prohibitively expensive, while adapting on randomly sampled spans introduces severe noise. 
Because most spans in a long context are irrelevant to the specific question, training on them may even degrade the base model's performance.
Our preliminary study shows that TTT is highly sensitive to training-span quality: on LongBench-v2, TTT on randomly sampled spans hurts performance, whereas TTT on oracle spans substantially improves it.
Motivated by this, we propose a simple method, \ourmethod{} (S-TTT): before adaptation, the model identifies the evidence spans it should learn from, and the standard language-modeling training objective is applied only to those selected spans.
On two challenging long-context reasoning benchmarks, LongBench-v2 and LongBench-Pro, S-TTT improves accuracy for both Qwen3-4B-Thinking-2507 and Llama-3.1-8B-Instruct, achieving up to a 15\% relative improvement.}

\begin{document}
\maketitle

\section{Introduction}

Long-context capability has become a central requirement for modern language models. Recent models support context windows of hundreds of thousands of tokens, enabling them to process long inputs in a single prompt~\citep{peng2024yarn,chen2024longlora}. Despite this progress, a larger window does not by itself ensure that the model can use long inputs effectively. As context length grows, accuracy often degrades, and models struggle to keep the most relevant evidence accessible throughout reasoning and decoding \citep{liu2024lost,hsieh2024ruler}. This suggests that the bottleneck in long-context reasoning is not merely fitting more tokens into the prompt, but ensuring that the model can identify and use the evidence relevant to the question.
Test-time training (TTT)~\citep{sun2020ttt,liu2021tttpp,hardt2024nearest,akyurek2024ttt,tandon2025end,zhang2025test,feng2026inplace} has emerged as a promising solution. Instead of answering with a fixed model, TTT treats the test input itself as a training example, adapts the model weights for that specific instance, and uses the adapted weights to generate the answer.
For long-context tasks, this is especially appealing because adaptation can turn instance-specific evidence in the context into parameter updates, making it easier to use during subsequent generation~\citep{bansal2025qttt,chen2025perk}.

However, a key challenge in applying TTT to long contexts is determining \emph{what} data to train on---an important dimension that remains largely underexplored. 
Existing approaches commonly rely on either full-context adaptation~\citep{tandon2025end,zhang2025test} or randomly sampled training spans\citep{bansal2025qttt}, both of which suffer from noisy signals.
Not only is performing TTT on the full context computationally expensive, but it also overwhelms the adaptation process with distractors, as the vast majority of a long context is usually irrelevant to the specific query. 
A cheaper alternative is to train on randomly sampled spans. While this mitigates the computational cost, it may amplify the noise: random sampling frequently misses the relevant evidence, causing the model to adapt primarily on distractors.

This suggests that the central bottleneck of long-context TTT is not the adaptation mechanism itself, but rather \emph{test-time training-data quality}. We empirically demonstrate this sensitivity through a preliminary diagnostic: on LongBench-v2~\citep{bai2025longbenchv2}, TTT on random spans slightly degrades performance relative to standard base model inference, whereas training on answer-aware oracle spans annotated by GPT-5.5 yields substantial improvements. This demonstrates that the effectiveness of TTT depends critically on the signal-to-noise ratio of the training tokens.

Motivated by this insight, we propose a simple solution, \ourmethod{} (S-TTT). Rather than processing the entire context or sampling spans blindly, S-TTT leverages the LLM itself as a test-time data selector.
We prompt the model to mark verbatim spans in the context that are likely to support the question. We then adapt the model on the selected spans with a next-token-prediction objective and generate the final answer from the full context. 
As such, S-TTT leaves the training objective, model architecture, and final decoding procedure unchanged; it optimizes only the test-time tokens used for adaptation.
On two challenging long-context reasoning benchmarks, LongBench-v2~\citep{bai2025longbenchv2} and LongBench-Pro~\citep{longbenchpro}, using Qwen3-4B-Thinking-2507~\citep{qwen3} and Llama-3.1-8B-Instruct~\citep{llama3} models, S-TTT consistently improves long-context performance and outperforms strong TTT baselines.

Our contributions are:
\begin{enumerate}
  \item We identify \emph{training-data quality} as a critical yet underexplored bottleneck for long-context TTT. We empirically demonstrate that adapting on noisy context can degrade performance, whereas high-quality evidence spans lead to substantial gains.
  \item We propose \ourmethod{} (S-TTT), a simple and effective framework that uses the LLM itself to select question-relevant evidence spans for test-time training, avoiding the expensive computational cost of full-context training and mitigating the severe noise of random span sampling.
  \item We evaluate S-TTT on two challenging long-context reasoning benchmarks LongBench-v2 and LongBench-Pro using Qwen3-4B-Thinking and Llama-3.1-8B-Instruct models. S-TTT consistently improves long-context performance and outperforms various strong TTT baselines.
\end{enumerate}

\section{Method}
\label{sec:method}

\subsection{Preliminary analysis}
\label{subsec:prelim}
Test-time training has been used to improve LLMs long-context performance by adapting the model to the specific context observed at test time. For long inputs, however, directly training on the full context is expensive, and a naive alternative is to train on short spans sampled uniformly from the context. This reduces the compute but has a cost: in a long document, most uniformly sampled spans are irrelevant to the question. As a result, TTT on randomly sampled spans may adapt the model to distractors rather than evidence.

\begin{table}[h]
\centering
\small
\begin{tabular}{lc}
\toprule
\textbf{Method} & \textbf{LongBench-v2} \\
\midrule
Base Model & 40.4 \\
Random Span TTT & 38.9 \\
Oracle Span TTT & \textbf{45.9} \\
\bottomrule
\end{tabular}
\caption{Test-time training is sensitive to training-token quality. Training Qwen3-4B-Thinking-2507 on random span tokens does not lead to improvement; instead, it hurts performance.}
\label{tab:prelim_token_quality}
\end{table}

Table~\ref{tab:prelim_token_quality} shows a diagnostic experiment on LongBench-v2. The base model Qwen3-4B-Thinking-2507 reaches $40.4\%$ accuracy without fine-tuning. After adapting the model via TTT on uniformly sampled spans, accuracy drops to $38.9\%$, indicating that TTT does not guarantee an improvement when the training tokens are noisy. In contrast, when the training spans are oracle spans annotated by GPT-5.5 with access to the ground-truth answer, the same TTT procedure achieves $45.9\%$ accuracy. Notably, we explicitly control the length of oracle spans to be comparable to that of the random spans, therefore, the number of training tokens is not the factor affecting the performance. This gap isolates the role of the training data quality: TTT can help, but only when the tokens used contain useful evidence.

This motivates our core view: the central bottleneck in long-context TTT is not only how to adapt the model, but also what to adapt on. 
High-quality spans provide a much stronger training signal, however, relying on an external oracle is not a practical solution. We therefore ask whether the model can identify the effective test-time training tokens by itself.

\begin{figure}[t]
    \centering
    \includegraphics[width=\linewidth]{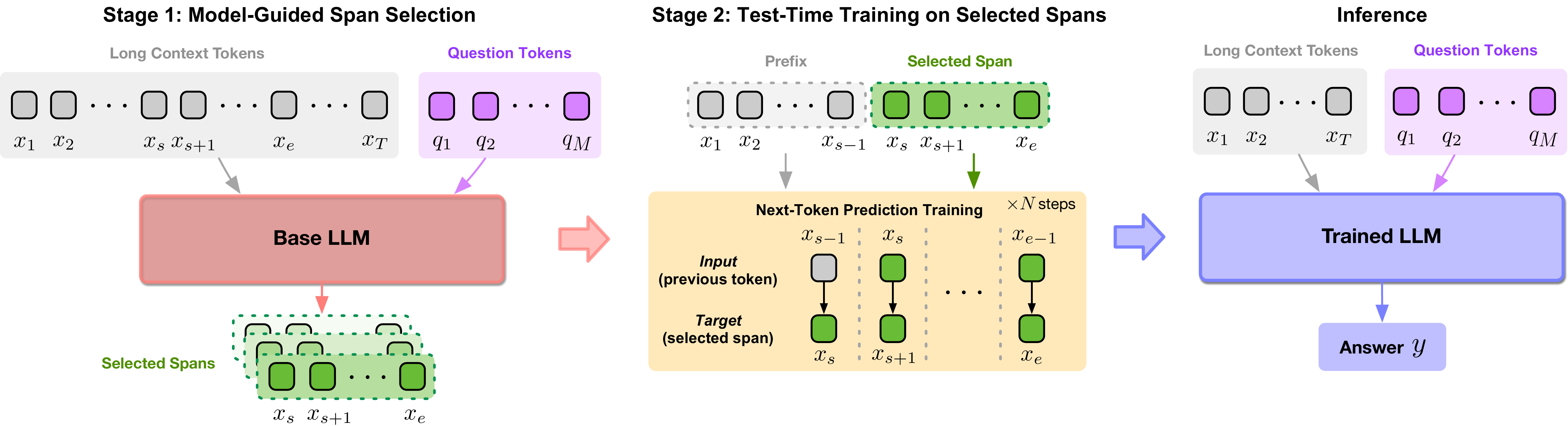}
    \caption{Overview of \ourmethod{}.
  In Stage 1, the base LLM reads the long context and question, and identifies question-relevant spans from the context.
  In Stage 2, these selected spans are used for TTT.
  At inference time, the adapted model generates the answer conditioning on the original full context and the question.}
    \label{fig:main_figure}
\end{figure}

\subsection{\ourmethod{}}
The preliminary analysis suggests that test-time training is effective only when the training tokens provide useful information for the current test instance. Based on this observation, we introduce \ourmethod{} (S-TTT), in which the model first identifies question-relevant evidence from the context and then adapts itself on the selected evidence.
Specifically, given a context
$x = (x_1,\ldots,x_T)$, a question $q$, and a base model with parameters
$\theta$, S-TTT consists of two stages:

\textbf{Stage 1: Model-guided span selection.}
We first ask the model to identify the parts of the context that are most
relevant to answering $q$. Concretely, the model reads the full context and
question and returns a set of verbatim supporting spans,
\[
\mathcal{S}(x,q)
=
\left\{
x_{s_j:e_j}
\right\}_{j=1}^{M},
\]
where each interval $[s_j,e_j]$ corresponds to a contiguous span copied from
the original context. This selection step relies on the model's own relevance judgment to construct
instance-specific training data. The purpose of this stage is not to replace
the original context at generation time, but to identify the subset of tokens
that provides the most useful adaptation signal, which may otherwise be
buried among a large amount of irrelevant information in the full context.

\textbf{Stage 2: Test-time training on selected spans.}
Starting from a fresh copy of the base model $\theta' \leftarrow \theta$,
we perform next-token prediction on the selected spans. For a
selected span $x_{s_j:e_j}$, the training objective is
\[
\mathcal{L}_{\mathrm{TTT}}(\theta')
=
-\sum_{i=s_j}^{e_j}
\log p_{\theta'}\!\left(x_i \mid x_{<i}\right).
\]
Across adaptation steps, we cycle through the valid spans in
$\mathcal{S}(x,q)$ and update $\theta'$ using the training objective above. The model is encouraged to internalize information that is likely to be
useful for answering the current question identified by itself, rather than arbitrary content from
the long context.

After adaptation, the updated model generates the answer conditioned on the original
full context and question:
\[
y \sim p_{\theta'}(\,\cdot \mid x_{1:T},q).
\]
The full context remains available during generation, so span selection
determines only the test-time training data and does not remove potentially
useful information from the final input. Once the instance is completed,
$\theta'$ is discarded and the next instance begins from the original
parameters $\theta$. 
A per-instance loop is described in Algorithm~\ref{alg:loop}.

\begin{algorithm}[t]

\caption{\ourmethod{}}
\label{alg:loop}
\begin{algorithmic}[1]
\Require Model $\theta$, context $x_{1:T}$, question $q$, steps $N$, span length $k$, learning rate $\eta$
\State Initialize a fresh model $\theta'$ from $\theta$
\State $\mathcal{S}\gets$ spans in $x_{1:T}$ annotated by $\theta$ relevant to $q$
\If{$\mathcal{S}=\emptyset$}
    \State $\mathcal{S}\gets$ random spans sampled from $x_{1:T}$
\EndIf
\For{$n=1,\dots,N$}
    \State Choose span $x_{s_{j}:e_{j}}$ from $\mathcal{S}$
    \State $\mathcal{L}_{\mathrm{TTT}}\gets-\sum_{i=s_{j}}^{e_{j}}\log p_{\theta'}(x_i\mid x_{<i})$
    \State Update $\theta' \gets \theta' - \eta \nabla \mathcal{L}_{\text{TTT}}$
\EndFor 
\State \Return answer  $y \sim p_{\theta'}(\, \cdot \mid x_{1:T},q)$

\end{algorithmic}
\end{algorithm}

\section{Experimental Results}
\label{sec:setup}

\subsection{Setup}
\label{subsec:setup}

\paragraph{Models and benchmarks.}
We evaluate two base models:
\textbf{Qwen3-4B-Thinking-2507}~\citep{qwen3} and
\textbf{Llama-3.1-8B-Instruct}~\citep{llama3}. We conduct experiments on two challenging long-context benchmarks.
\textbf{LongBench-v2}~\citep{bai2025longbenchv2} is a four-way
multiple-choice benchmark covering diverse long-context reasoning tasks and
is evaluated using answer accuracy.
\textbf{LongBench-Pro}~\citep{longbenchpro} evaluates a broader set of long-context capabilities and has English and Chinese subsets, we use its English subset as our evaluation set and apply its official
evaluation pipeline for scoring. We use the Qwen3 tokenizer to measure context length and keep examples whose contexts contain at most $128$k tokens.

\paragraph{Compared methods.}
We compare the following settings:

\begin{itemize}
    \item \textbf{Base Model.}
    The base model directly generates an answer conditioned on the full
    context and question, without any parameter updates.

    \item \textbf{LongLLMLingua.}
    LongLLMLingua~\citep{jiang2024longllmlingua} is a prompt compression method. The base model first compresses the full context into a shorter one conditioned on the question, and then answers the question using the compressed context. We set the compressed-context budget to be $4{,}096$ tokens.

    \item \textbf{qTTT.} qTTT~\citep{bansal2025qttt}  is an efficiency-oriented TTT method that adapts on uniformly sampled random spans. It first runs a single forward pass over the full context to build the KV cache, then keeps the cache frozen and updates only the query-projection parameters, avoiding recomputation of the full-context KV at every adaptation step.
    
    \item \textbf{QRHead Span TTT.}
    Following QRHead~\citep{zhang2025qrhead},  we identify query-relevant attention heads on a retrieval set BEIR~\citep{thakur2021beir} by scoring each head's query-to-context attention as a retriever and the $16$ highest-scoring heads are kept as QRHeads. At test time, we run a single forward pass over the full context to obtain QRHead attention scores, aggregate them over each $512$-token candidate span to obtain span-level scores, and select the $8$ highest-scoring spans for TTT.

    \item \textbf{Random Span TTT.}
    We randomly sample $8$ spans from the context, each containing
    $512$ tokens, and use them for TTT. Random Span TTT differs from  qTTT in updating with a non-frozen KV cache.

    \item \textbf{Full Context TTT.}
    We partition the full context into $N$ contiguous chunks, where $N$ is the number of adaptation steps. At each step, we perform one step TTT update on one chunk.
    
    \item \textbf{\ourmethod{} (ours).}
    We ask the model to identify at most $8$ spans in the context that are relevant to the
    question and then perform TTT on the selected spans. If the model fails to output valid spans, it falls back to using uniformly sampled spans. 
    We report fallback rates in Appendix~\ref{sec:annotation_coverage}.
    
\end{itemize}

For all TTT methods, the final answer is generated conditioned on the full context rather than the selected spans.
We use LoRA for parameter-efficient test-time adaptation and perform
$16$ gradient-update steps for each test instance. More details can be found in Appendix \ref{sec:implementation_details}.

\paragraph{Evaluation.}
For each test instance, we sample $4$ responses and evaluate each
response using the corresponding benchmark evaluator. We report the mean scores for the four samples. We sample responses with a temperature of $0.6$ and a top-$p$ of $0.95$. The maximum generation length is set to $32{,}768$
tokens for Qwen3-4B-Thinking-2507 and $10{,}240$ tokens for Llama-3.1-8B-Instruct. Prompt templates can be found in Appendix \ref{sec:prompt}.

\subsection{Results}
\label{sec:results}

\begin{table*}[t]
\centering
\caption{
Results on LongBench-v2 and LongBench-Pro across different context-length buckets. The best result for each model and evaluation setting is shown in \textbf{bold}. \textsuperscript{$\dagger$}QRHead Span TTT is not directly comparable with the other TTT methods as it requires additional information to identify retrieval heads.
}
\label{tab:main_results}
\small
\setlength{\tabcolsep}{5pt}
\renewcommand{\arraystretch}{1.15}

\begin{tabularx}{\textwidth}{
    >{\centering\arraybackslash}p{2.3cm}
    >{\raggedright\arraybackslash}p{3.0cm}
    >{\centering\arraybackslash}X
    >{\centering\arraybackslash}X
    >{\centering\arraybackslash}X
    >{\centering\arraybackslash}X
}
\toprule
\multirow{2}{*}{\textbf{Model}}
& \multirow{2}{*}{\textbf{Method}}
& \multicolumn{2}{c}{\textbf{LongBench-v2}}
& \multicolumn{2}{c}{\textbf{LongBench-Pro}} \\
\cmidrule(lr){3-4}
\cmidrule(lr){5-6}
&
& \textbf{$<64\mathrm{k}$}
& \textbf{$64\mathrm{k}$--$128\mathrm{k}$}
& \textbf{$<64\mathrm{k}$}
& \textbf{$64\mathrm{k}$--$128\mathrm{k}$} \\
\midrule

\multirow{7}{2.5cm}{\centering
    \textbf{\makecell{Qwen3-4B-\\Thinking-2507}}
}
& Base Model
& 46.7
& 30.7
& 55.1
& 41.6 \\

& LongLLMLingua
& 41.8
& 31.7
& 35.0
& 30.3 \\

& qTTT
& 44.7
& 34.0
& 56.6
& 41.5 \\

& QRHead Span TTT\textsuperscript{$\dagger$}
& 47.2
& 32.1
& \textbf{56.7}
& 40.8 \\

& Random Span TTT
& 43.6
& 34.2
& 55.0
& 41.0 \\

& Full Context TTT
& 45.1
& 32.6
& 55.8
& 40.4 \\

& \ourmethod{}
& \textbf{47.7}
& \textbf{35.3}
& 56.2
& \textbf{42.0} \\

\midrule

\multirow{7}{2.5cm}{\centering
    \textbf{\makecell{Llama-3.1-\\8B-Instruct}}
}
& Base Model
& 36.9
& 26.3
& 28.2
& 19.4 \\

& LongLLMLingua
& 34.1
& 26.9
& 25.2
& 21.0 \\

& qTTT
& 35.7
& 27.5
& 29.7
& 19.3 \\

& QRHead Span TTT\textsuperscript{$\dagger$}
& 35.7
& 27.5
& 29.4
& 20.4 \\

& Random Span TTT
& 36.0
& 26.7
& 28.7
& 20.4 \\

& Full Context TTT
& 35.2
& 27.7
& 29.4
& 19.8 \\

& \ourmethod{}
& \textbf{38.4}
& \textbf{28.2}
& \textbf{29.9}
& \textbf{21.7} \\

\bottomrule
\end{tabularx}
\end{table*}

Table~\ref{tab:main_results} summarizes the main results. We highlight three observations. First, S-TTT consistently improves over the base model across models, benchmarks, and length buckets. Second, S-TTT consistently outperforms or is comparable to all the other TTT methods, showing that model-annotated training spans provide a more reliable adaptation signal than uniformly sampled spans or full context. Third, the gains are especially pronounced in the longer-context buckets, where irrelevant context is more abundant and training-token selection becomes more important.

\textbf{LongBench-v2.}
Using Qwen3-4B-Thinking-2507 as the base model, Random Span TTT degrades the $<64$k bucket, reducing accuracy from $46.7$ to $43.6$, whereas S-TTT improves it to $47.7$. In the $64$k--$128$k bucket,  all TTT baselines help, but S-TTT leads to the strongest score, reaching $35.3$. QRHead Span TTT is competitive in the shorter bucket, but drops behind in the longer bucket, suggesting that attention-based span scores are less stable as the context grows. Other baselines are less consistent: LongLLMLingua often underperforms the base model, and Full Context TTT remains below S-TTT in every setting.

The trend also transfers to Llama-3.1-8B-Instruct. S-TTT gives the best LongBench-v2 scores in both buckets, improving the base model from $36.9$ to $38.4$ and from $26.3$ to $28.2$, respectively. This indicates that the gain from S-TTT is model-agnostic.

\textbf{LongBench-Pro.}
With Qwen3-4B-Thinking-2507, S-TTT improves over the base model in both length buckets. In the shorter bucket, qTTT and QRHead Span TTT are slightly higher than S-TTT. In the longer bucket, however, S-TTT is the strongest method, reaching $42.0$ and outperforming all TTT baselines. This mirrors the LongBench-v2 trend: model-annotated spans become more valuable when the context is longer and noisier.

With Llama-3.1-8B-Instruct, S-TTT again gives the best LongBench-Pro scores in both buckets, improving the base model from $28.2$ to $29.9$ and from $19.4$ to $21.7$, respectively. Overall, these results support our main hypothesis that training-token quality is a central bottleneck for long-context TTT.

\section{Analysis}
\label{sec:analysis}

We further analyze why and when S-TTT works. We ask three questions: 
(1) whether question-conditioned model annotation is more effective than annotation-free intrinsic span scores, 
(2) how adaptation on selected spans changes the model's attention to the relevant evidence, and 
(3) how the end-to-end overhead of S-TTT scales with context length.

\subsection{Span selection strategies}

The main results show that span selection matters. We next ask whether explicit model annotation is necessary, or whether simpler annotation-free signals can select useful training spans. A natural alternative is to use intrinsic model statistics, selecting spans that are difficult to predict or induce high uncertainty in the next-token distribution.

We compare against two intrinsic selectors. The perplexity selector ranks each $512$-token window by mean negative log-likelihood, while the entropy selector ranks each window by mean predictive entropy. Each method then performs TTT on the top $8$ highest-scoring spans. We keep the training configuration fixed across all methods, changing only how the training spans are selected.

\begin{table}[h]
\centering
\small
\begin{tabular}{lcc}
\toprule
\textbf{Span selector} & \textbf{$<$64k} & \textbf{64--128k} \\
\midrule
Model annotation & \textbf{47.7} & \textbf{35.3} \\
Perplexity score & 46.7 & 31.9 \\
Entropy score & 45.1 & 33.0 \\
\bottomrule
\end{tabular}
\caption{Model-annotated spans outperform intrinsic metric-selected spans on LongBench-v2 with Qwen3-4B-Thinking-2507.}
\label{tab:intrinsic_span_selectors}
\end{table}

Table~\ref{tab:intrinsic_span_selectors} shows that model-annotated spans perform best in both length buckets, indicating that model intrinsic metrics are not the best choice for selecting useful TTT data. The gap is small below $64$k for perplexity-selected spans but becomes much larger in the $64$k--$128$k bucket. This is the regime where the context contains more distractors and where selecting question-relevant evidence becomes most important.

These results suggest that useful TTT spans are not simply the spans that are surprising or uncertain under the language model. High-perplexity or high-entropy text may be difficult to predict for many reasons unrelated to the question, such as formatting, rare entities, or local distribution shift. In contrast, model annotation conditions span selection on the question, allowing it to better target evidence that can improve the final answer.

\begin{figure}[h]
\centering
\includegraphics[width=\linewidth]{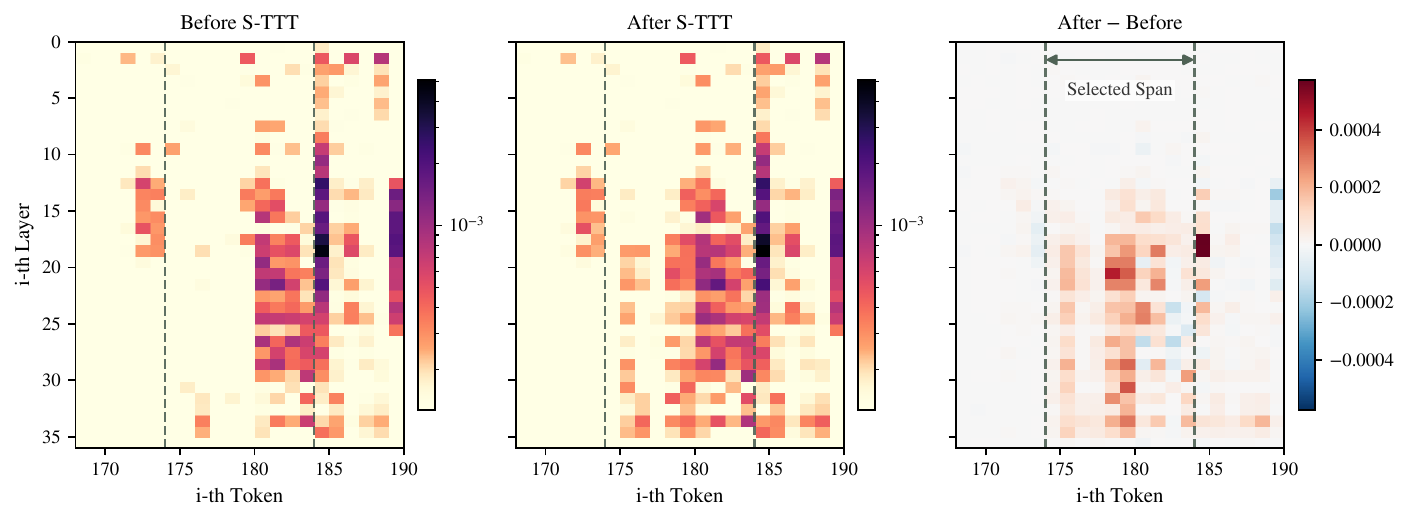}
\caption{A concrete example of question-and-answer-to-context attention before and after S-TTT. Rows are layers and columns are context positions around the model-annotated span. The dashed vertical lines mark the selected training tokens. After S-TTT, attention to the selected span increases, while the change outside the span remains small.}
\label{fig:attention_spanzoom_case}
\end{figure}

\subsection{Case study}

We next visualize how S-TTT changes the model's use of the selected span. We compare question-and-answer-to-context attention before and after S-TTT, averaging over all heads and plotting the attention by layer.
Figure~\ref{fig:attention_spanzoom_case} shows one such example. Before adaptation, the model already assigns some attention to the annotated evidence span, but the mass is sparse and uneven across layers. After training on that span, attention becomes stronger and more continuous around the selected tokens, especially in the middle layers. The difference panel shows that this change is localized: the warm region aligns with the training span, while most neighboring positions remain close to zero. This qualitative example suggests one mechanism behind S-TTT: adaptation on selected evidence induces a localized shift in attention toward tokens that are relevant to the current question. More visualized examples can be found in Appendix \ref{sec:attention_examples}.

\begin{figure}[t]
\centering
\includegraphics[width=1.0\linewidth]{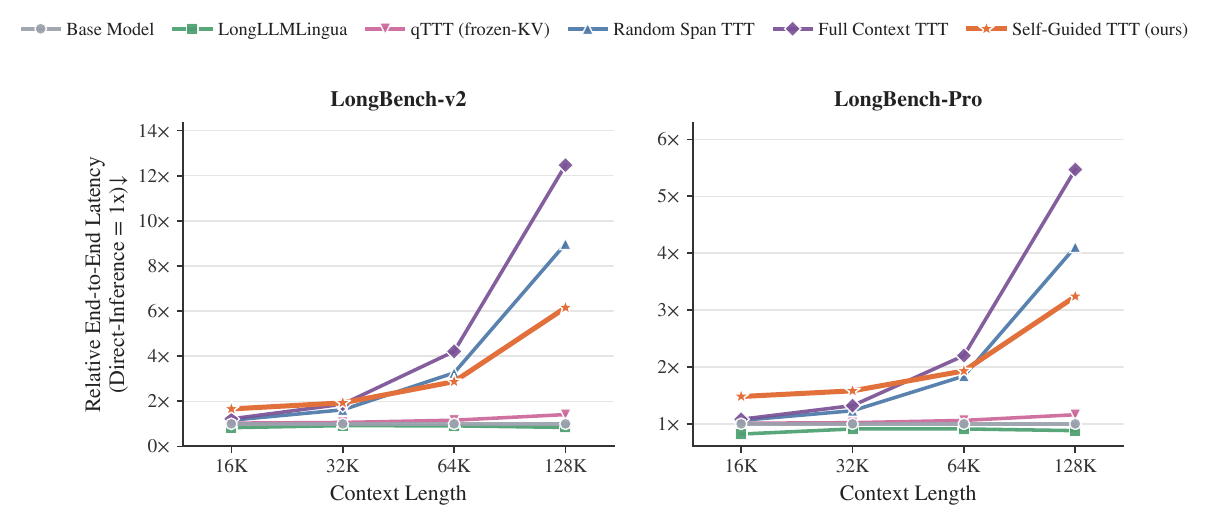}
\caption{End-to-end latency normalized by full-context inference as context length increases using Qwen3-4B-Thinking-2507. S-TTT incurs a higher latency in the beginning, but it becomes cheaper than other non-frozen KV cache TTT methods at longer context.}
\label{fig:latency_vs_context}
\end{figure}

\subsection{Efficiency analysis}

TTT methods introduce extra overhead over direct inference because they add an adaptation stage before generation.
We use pytorch FSDP~\citep{paszke2019pytorch} for training and vLLM~\citep{vllm} for inference. All measurements are conducted on a single NVIDIA H200 GPU. Figure~\ref{fig:latency_vs_context} reports measured end-to-end latency normalized by full-context inference using Qwen3-4B-Thinking-2507. S-TTT has a higher latency in the beginning when the context length is relatively short. The crossover happens at longer context: S-TTT becomes cheaper than Full Context TTT from $64$k onward on both benchmarks, and cheaper than Random Span TTT at $64$k on LongBench-v2 and comparable on LongBench-Pro. Notably, at $128$k context length, S-TTT has the lowest latency among the non-frozen-KV TTT methods. This is expected because Full Context TTT will incur significant overhead when the context length scales up as it trains on the entire input. Random Span TTT samples spans uniformly across the full context, which leads to a larger average effective training window of $0.50C$, where $C$ is the context length. In contrast, model-annotated spans are more localized, resulting in shorter effective training windows on average ($0.39C$ on LongBench-v2 and $0.37C$ on LongBench-Pro). The annotation cost dominates at shorter lengths, but quickly becomes smaller than the saved adaptation cost at long context.

\section{Related Work}

\textbf{Test-Time Training.} TTT adapts model parameters to a single test input before prediction, using supervision derived from the input itself rather than from new labels \citep{sun2020ttt}. Earlier work studies when self-supervised TTT helps or fails under distribution shift \citep{liu2021tttpp}, and nearest-neighbor TTT adapts LLMs using retrieved examples at inference time \citep{hardt2024nearest}. More recent LLM work shows that per-instance adaptation can improve reasoning when the test input contains useful self-supervision \citep{akyurek2024ttt}. For long-context tasks, \citet{bansal2025qttt} show that TTT can be a more effective use of inference-time compute than simply generating more reasoning tokens. Related long-context TTT work also explores parameter-efficient adaptation for reasoning over long inputs \citep{chen2025perk}. These works primarily study how to perform adaptation efficiently. In this work, we study what tokens the model should be trained on at test time. S-TTT shows that selecting the right spans is a key component of effective long-context TTT.

\textbf{Long-Context LLMs.} Modern LLMs increasingly support very long context windows, but a longer window does not guarantee reliable use of the information inside it. Models remain sensitive to evidence position, often degrading when relevant content appears in the middle of a long input \citep{liu2024lost}, and long-context benchmarks such as LongBench, LongBench-v2, LongBench-Pro, ZeroSCROLLS, RULER, and HELMET make these failures visible across multi-document QA, code, dialogue, structured reasoning, recall, and long in-context learning tasks \citep{bai2024longbench,bai2025longbenchv2,longbenchpro,shaham2023zeroscrolls,hsieh2024ruler,yen2025helmet}. A broad line of work addresses long-context limitations by extending usable context windows \citep{peng2024yarn,chen2024longlora}, improving prefill or attention efficiency \citep{jiang2024minference}, compressing prompts \citep{jiang2024longllmlingua}, retrieving external evidence \citep{lewis2020rag, zhang2025qrhead}, or analyzing and steering attention behavior at inference time \citep{wu2024retrievalhead,zhang2025qrhead,ye_dysco_2026}. These approaches largely aim to help the model condition on the right evidence or process long inputs more efficiently.
We instead approach long-context reasoning from the perspective of test-time training: rather than compressing the long context or intervening in the decoding procedure, S-TTT only requires the model to first select relevant spans from the context before TTT. This keeps the model architecture and decoding algorithm unchanged, avoiding complex interventions that are often infeasible in modern inference engines, while yielding consistent gains.

\section{Conclusion}

We propose \ourmethod{} (S-TTT), a simple test-time adaptation framework for long-context LLMs that uses the model itself to select question-relevant evidence spans for training. Instead of adapting on the full context or on randomly sampled spans, S-TTT first identifies supporting spans from the input context, adapts the model only on those selected spans, and then generates the final answer using the original full context. Our results on LongBench-v2 and LongBench-Pro show that S-TTT consistently improves over TTT on random span across Qwen3 and Llama-3.1 models, while remaining cheaper than other TTT variants at long context. Empirical results demonstrate that the effectiveness of long-context TTT depends critically on the quality of the test-time training tokens. Overall, S-TTT provides a simple yet effective framework for long-context test-time training, highlighting training-token selection as a promising direction for solving long-context tasks with TTT. We discuss future directions in Appendix \ref{sec:future_work}.

% \clearpage
\bibliographystyle{assets/plainnat}
\bibliography{custom}

\clearpage
\beginappendix
\section{Implementation Details}
\label{sec:implementation_details}

We use LoRA~\citep{hu2022lora} for parameter-efficient test-time training. Following qTTT~\citep{bansal2025qttt}, we apply LoRA only to the query projection layers, with rank $r=16$ and scaling parameter $\alpha=32$. We optimize the LoRA parameters using AdamW with a $0.01$ weight decay. For each method, we sweep the learning rate over $\{3\times10^{-5},\,1\times10^{-4},\,3\times10^{-4}\}$ on a small validation set and select the best one for testing. For span annotation, LongBench-v2, which consists of multiple-choice questions, we append the answer choices to the question when prompting the model to annotate relevant spans. For LongBench-Pro, which contains open-ended questions, span annotation is performed using only the context and the question.

\section{Annotation Coverage}
\label{sec:annotation_coverage}

\begin{table}[h]
\centering
\caption{Model annotation coverage on LongBench-v2 and LongBench-Pro. ``Fallback'' means the model fails to produce valid verbatim spans.}
\small
\begin{tabular}{llc}
\toprule
\textbf{Model} & \textbf{Benchmark}  & \textbf{Fallback} \\
\midrule
Qwen3-4B-Thinking-2507 & LongBench-v2  & 8.2\% \\
Qwen3-4B-Thinking-2507 & LongBench-Pro  & 21.5\% \\
Llama-3.1-8B-Instruct & LongBench-v2  & 6.9\%  \\
Llama-3.1-8B-Instruct & LongBench-Pro & 39.9\% \\
\bottomrule
\end{tabular}
\label{tab:annotation_coverage}
\end{table}

Table~\ref{tab:annotation_coverage} reports how often the model produces valid verbatim spans. Fallback instances use random spans, so they are equivalent to Random Span TTT for those cases. The fallback rate is low on LongBench-v2 for both base models, indicating that most instances receive genuine model-selected training spans. However, the fallback rate becomes higher on LongBench-Pro, especially for Llama-3.1-8B-Instruct, suggesting that self-annotation is more difficult on the open-ended benchmark.

\section{Prompts}
\label{sec:prompt}
\begin{table*}[h]
\centering
\small
\caption{
Qwen3-4B-Thinking-2507 prompt template for LongBench-v2.}
\begin{tabular}{@{}p{0.96\textwidth}@{}}
\toprule
\ttfamily
\detokenize{<|im_start|>system}\par
\detokenize{You are a helpful assistant. Read the context and answer the question.<|im_end|>}\par
\detokenize{<|im_start|>user}\par
\detokenize{{CONTEXT}}\par
\mbox{}\par
\detokenize{{QUESTION}}\par
\detokenize{Pick from the following options:}\par
\detokenize{{CHOICES}}\par
\detokenize{Please show your choice in the answer field with only the choice letter,}\par
\detokenize{e.g., "answer": "C".<|im_end|>}\par
\detokenize{<|im_start|>assistant}\par
\detokenize{<think>}
\\
\bottomrule
\end{tabular}
\label{tab:qwen-longbench-v2-prompt}
\end{table*}

\begin{table*}[h]
\centering
\caption{
Qwen3-4B-Thinking-2507 prompt template for LongBench-Pro.
}
\label{tab:qwen-longbench-pro-prompt}
\small
\begin{tabular}{@{}p{0.96\textwidth}@{}}
\toprule
\ttfamily
\detokenize{<|im_start|>system}\par
\detokenize{You are a helpful assistant. Read the context and answer the question.<|im_end|>}\par
\detokenize{<|im_start|>user}\par
\detokenize{{CONTEXT}}\par
\mbox{}\par
\detokenize{{QUESTION}<|im_end|>}\par
\detokenize{<|im_start|>assistant}\par
\detokenize{<think>}
\\
\bottomrule
\end{tabular}
\end{table*}
\begin{table*}[h]
\centering
\caption{
Llama-3.1-8B-Instruct prompt template for LongBench-v2.
}
\label{tab:llama-longbench-v2-prompt}
\small
\begin{tabular}{@{}p{0.96\textwidth}@{}}
\toprule
\ttfamily
\detokenize{<|begin_of_text|><|start_header_id|>system<|end_header_id|>}\par
\mbox{}\par
\detokenize{Cutting Knowledge Date: December 2023}\par
\detokenize{Today Date: 26 Jul 2024}\par
\mbox{}\par
\detokenize{<|eot_id|><|start_header_id|>user<|end_header_id|>}\par
\mbox{}\par
\detokenize{Please read the following text and answer the question below.}\par
\mbox{}\par
\detokenize{<text>}\par
\detokenize{{CONTEXT}}\par
\detokenize{</text>}\par
\mbox{}\par
\detokenize{What is the correct answer to this question: {QUESTION}}\par
\detokenize{Choices:}\par
\detokenize{{CHOICES}}\par
\mbox{}\par
\detokenize{Let's think step by step. After thinking, choose a single, most likely}\par
\detokenize{answer. Output your final answer follows: "The correct answer is}\par
\detokenize{(insert choice here)".<|eot_id|><|start_header_id|>assistant<|end_header_id|>}
\\
\bottomrule
\end{tabular}
\end{table*}
\begin{table*}[h]
\centering
\caption{
Llama-3.1-8B-Instruct prompt template for LongBench-Pro.
}
\label{tab:llama-longbench-pro-prompt}
\small
\begin{tabular}{@{}p{0.96\textwidth}@{}}
\toprule
\ttfamily
\detokenize{<|begin_of_text|><|start_header_id|>system<|end_header_id|>}\par
\mbox{}\par
\detokenize{Cutting Knowledge Date: December 2023}\par
\detokenize{Today Date: 26 Jul 2024}\par
\mbox{}\par
\detokenize{You are a helpful assistant. Read the context and answer the}\par
\detokenize{question.<|eot_id|><|start_header_id|>user<|end_header_id|>}\par
\mbox{}\par
\detokenize{{CONTEXT}}\par
\mbox{}\par
\detokenize{{QUESTION}<|eot_id|><|start_header_id|>assistant<|end_header_id|>}
\\
\bottomrule
\end{tabular}
\end{table*}

\FloatBarrier

\section{Qualitative Examples}
\label{sec:attention_examples}

\begin{figure}[h]
\centering
\includegraphics[width=\linewidth]{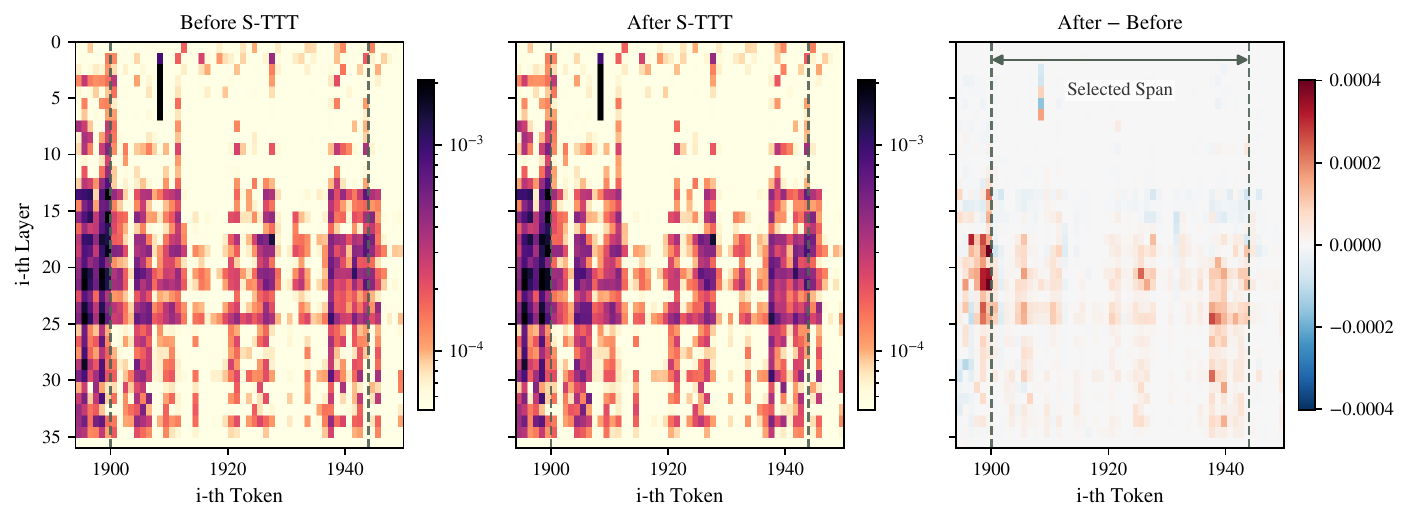}
\caption{Example of span attention before and after S-TTT.}
\end{figure}

\begin{figure}[h]
\centering
\includegraphics[width=\linewidth]{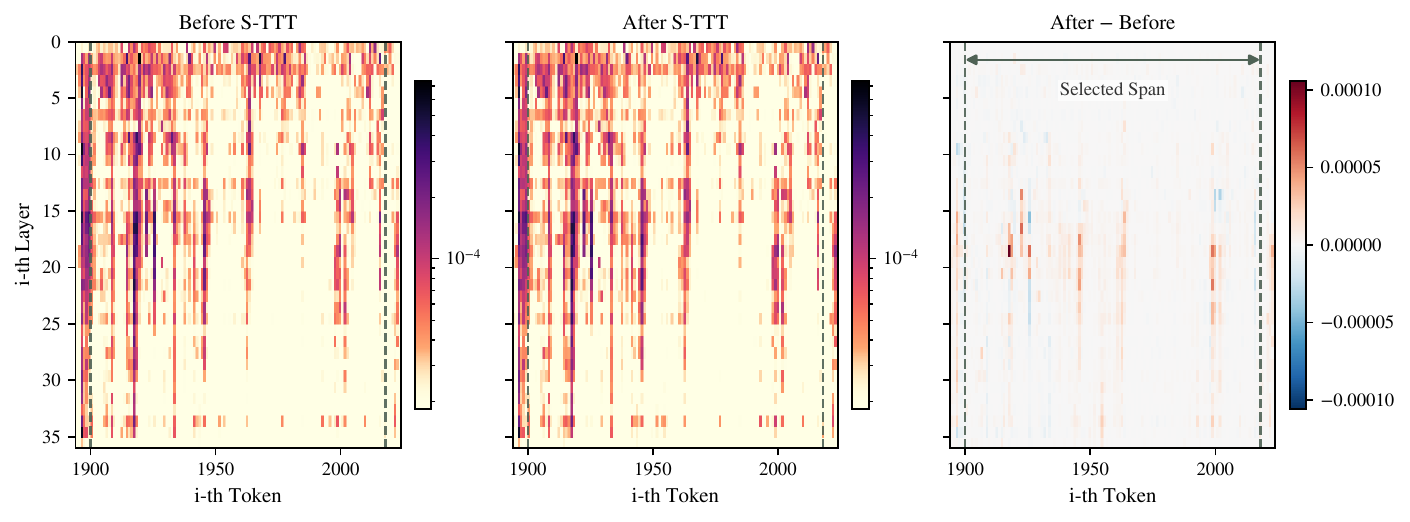}
\caption{Example of span attention before and after S-TTT.}
\end{figure}

\begin{figure}[h]
\centering
\includegraphics[width=\linewidth]{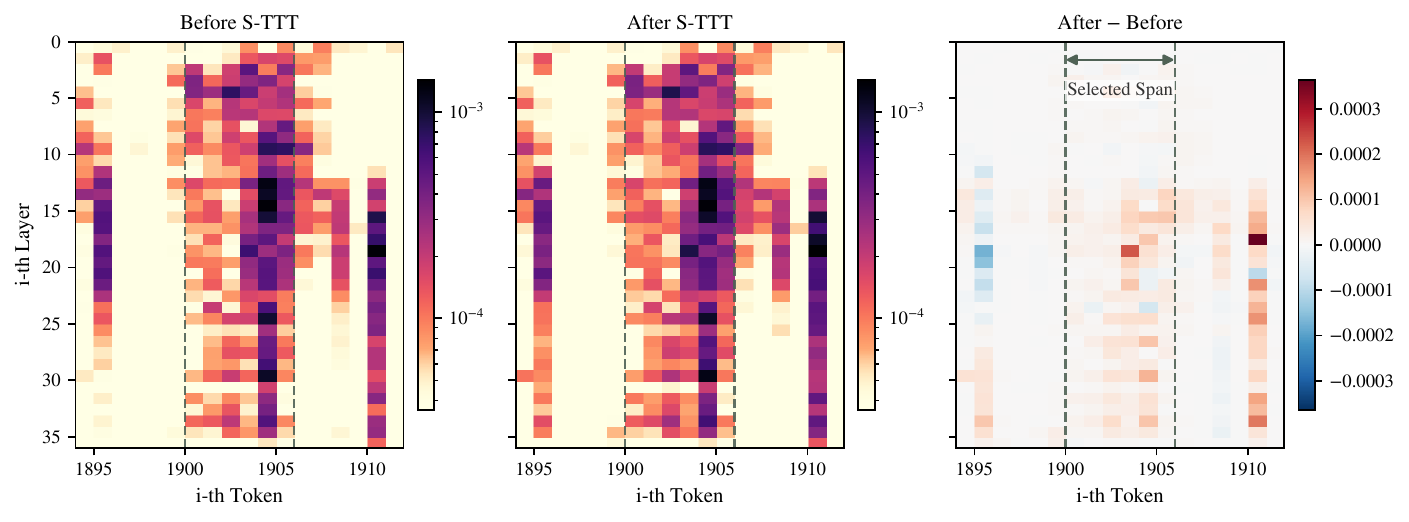}
\caption{Example of span attention before and after S-TTT.}
\end{figure}

\section{Future Directions}
\label{sec:future_work}

TTT opens up opportunities for adapting LLMs in realistic production settings. In many applications, users may upload a long document, such as a financial report, legal contract, or book, and then ask multiple questions about it. Unlike methods that require architectural changes or specialized attention mechanisms, TTT relies on standard gradient-based adaptation and on-the-fly weight updates. This makes it compatible with modern training and serving infrastructure, where each conversation session could maintain its own lightweight adapted weights for multi-turn use, personalization, or document-specific specialization.
However, the largest bottleneck is latency: even parameter-efficient TTT adds adaptation overhead before generation, which needs to be reduced. Addressing these challenges is an important direction for making TTT a practical framework for solving long-context tasks in real-world systems.

\end{document}